# The Effect of Epidemiological Cohort Creation on the Machine Learning Prediction of Homelessness and Police Interaction Outcomes Using Administrative Health Care Data

**Faezehsadat Shahidi[1], M. Ethan MacDonald[1-3], Dallas Seitz[3,4], Geoffrey Messier[1*]**

[1]Electrical and Software Engineering, Schulich School of Engineering, University of Calgary, Calgary, Alberta, Canada

[2]Biomedical Engineering, Schulich School of Engineering, University of Calgary, Calgary, Alberta, Canada

[3]Hotchkiss Brain Institute, University of Calgary, Calgary, Alberta, Canada

[4]Department of Psychiatry, Cumming School of Medicine, University of Calgary, Calgary, Alberta, Canada

**\* Correspondence:**
Corresponding Author
gmessier@ucalgary.ca

**Abstract**

**Background**: Mental illness can lead to adverse outcomes such as homelessness and police interaction and understanding of the events leading up to these adverse outcomes is important. Predictive models may help identify individuals at risk of such adverse outcomes. Using a fixed observation window cohort with logistic regression (LR) or machine learning (ML) models can result in lower performance when compared with adaptive and parcellated windows.

**Method**: An administrative healthcare dataset was used, comprising of 240,219 individuals in Calgary, Alberta, Canada who were diagnosed with addiction or mental health (AMH) between April 1, 2013, and March 31, 2018. The cohort was followed for 2 years to identify factors associated with homelessness and police interactions. To understand the benefit of flexible windows to predictive models, an alternative cohort was created. Then LR and ML models, including random forests (RF), and extreme gradient boosting (XGBoost) were compared in the two cohorts.

**Results**: Among 237,602 individuals, 0.8% (1,800) experienced first homelessness, while 0.32% (759) reported initial police interaction among 237,141 individuals. Male sex (AORs: H=1.51, P=2.52), substance disorder (AORs: H=3.70, P=2.83), psychiatrist visits (AORs: H=1.44, P=1.49), and drug abuse (AORs: H=2.67, P=1.83) were associated with initial homelessness (H) and police interaction (P). XGBoost showed superior performance using the flexible method (sensitivity =91%, AUC =90% for initial homelessness, and sensitivity =90%, AUC=89% for initial police interaction)

**Conclusion**: This study identified key features associated with initial homelessness and police interaction and demonstrated that flexible windows can improve predictive modeling.

# Introduction

Mental illnesses are known as a serious problem that have long-term negative effects on individuals' physical health (1), jobs (2), and society (2). According to the World Health Organization (WHO), around one in four people globally will experience mental health problems at some point in their lives. Homelessness is a significant problem worldwide. According to recent statistics, it is estimated that there are approximately 150 million homeless people globally at any given time (3). Statistics indicate that mentally ill persons are several times more likely to come into contact with police compared to the general population (4).

Based on a Government of Canada report, "Mental illness was experienced by 1 in 3 Canadians during their lifetime" (5,6). More than 235,000 Canadians experience homelessness each year (7), and 90% of them have mental disorders (8). Furthermore, each year approximately 5 million Canadians encounter police interaction (9) and between 15% to 40% (750,000 to 2 million) are known to have mental illness. There are serious potential consequences for those who have mental illnesses, including long delivery times for essential diagnosis and treatment, unneeded and harmful trauma, and criminalization of illness-related behavior (10). In Canada, the odds of police interaction for people with mental illnesses were 3.1 times higher than for people without mental illnesses (11).

Since people with mental illness are more likely to experience homelessness and encounters with the police, it is important to study the factors that are associated with these adverse outcomes. Additionally, many studies looked at the features associated with homelessness and police interaction, such as previous arrests (12) or previous use of emergency homeless shelters (13). However, they often neglect to examine the underlying causes that led to the initial occurrence (14). Specifically, these studies did not investigate the initial occurrence of these outcomes (14). Biomedical data, especially administrative data, are a potential great resource, however, working with administrative data has challenges (15), such as shift-invariance (different trajectory for each patient over time due to complex comorbidities), and scalability (large population with long duration longitudinal records are needed) (15).

To tackle the challenges of administrative data, most studies use an observation window length with fixed size by capturing the sequence of events and aggregating them into features by counting categories or averaging the numerical values (16,17). In this fixed window approach, the prediction window length contains the information (outcome event) the model is expected to use to make predictions, and the observation window length is the duration of observation before the prediction window, during which the model can gather features to make predictions (18). To illustrate the best size of the observation and prediction windows, some studies rely on the model performance by using various window lengths (18,19).

In the fixed window approach, participants with no entries during the observation window, with different times for the outcome window (cases where the outcomes occur before the prediction window), or with different lengths of historical data (cases where the historical events occur after the observation window) must be excluded (20) that may result in the loss of some patients with positive outcomes (illustrated in Figure 1 and noted by Gupta *et al*. (20)), which could affect the predictive models' performance. The existing literature lacks a well-defined strategy for creating cohorts using feature engineering to effectively retain more information.

LR models have certain limitations, such as managing highly correlated and continuous variables, as well as restrictions on the number of input variables (21). Although nonlinear ML models along with pre-processing methods generally exhibit superior predictive capabilities compared to linear ML models, like LR models (22,23), it was observed that when the fixed window method is employed, LR model can outperform nonlinear ML models (24). For example, Choi *et al.* (24). conducted several experiments on ML models and found that a linear model could outperform deep nonlinear neural network models using the fixed window method.

Two nonlinear ML models are going to be examined in this study. First is the RF model (25), an ML algorithm, that works based on the majority decisions of several 'decision trees' that were centered on randomly selected variables in a dataset (26). Second, XGBoost is appealing since, when compared to the other classifiers, its default option had a great average performance (27). XGBoost was based on a decision-tree ensemble algorithm that uses a gradient-boosting framework.

Some studies highlighted the limitations of the fixed window approach and emphasized the significance of selecting appropriate cohort creation methods to improve the performance of the predictive modeling. For example, Ng *et al.* (18) and Chen *et al.* (19) examined how well ML models perform in relation to the prediction window length and observation window length. These studies found that model performance improved as the prediction window length decreased and the observation window length increased. The authors observed that the model performance improved in relation to the density of patient encounters or the amount of available information. Also, Gupta *et al.* (20) used deep learning models to perform feature engineering methods rather than the fixed window approach and achieved superior results in prediction tasks.

To address the limitations of an epidemiology style cohort (fixed method), we propose the use of a flexible window method (shown in Figure 1 part (B)). By considering the length of individual records rather than using a fixed observation, the window size for everyone is determined based on their specific record length, allowing for a more adaptive approach in cohort creation and will capture more adverse mental health outcomes and enhance the effectiveness of predictive modeling.

This study aims to achieve two objectives. First, to employ a time-series administrative dataset in the field of AMH to create a cohort using a fixed observation window with a size of 5 years and a fixed outcome window with a size of 2 years (referred to as Primary Cohort), which is used to identify the factors associated with the initial occurrence of homelessness and police interaction among individuals diagnosed with AMH. Second, it sought to compare the performance of cohorts created using the flexible window method (where all data is included for a person up to the day of the first occurrence of outcome of interest, with a minimum history record of either 0, 30, 60, 90, 180, 360, or 720 that is referred to as Secondary Cohort) with those created using the fixed window method. Different pre-processing methods, as well as LR and ML models were compared to assess sensitivity and AUC and determine how cohort creation methods could enhance the predictive performance of AMH outcomes.

**Methods**

**Data Sources**

The data selected enabled a retrospective descriptive study using secondary data. The study protocol was reviewed and approved by the University of Calgary Ethics REB21-0070; Data Disclosure Agreements were developed with Alberta Health Services (AHS).

To gain insight into the local epidemiology and resource utilization implications of AMH diagnoses, a population-based study was conducted using administrative data collected between April 1, 2013, and March 31, 2020. The study population included individuals between the ages of 18 and 65 with AMH diagnoses as of April 1, 2018. The dataset contained information on all Calgary residents who were diagnosed with AMH between April 1, 2013, and March 31, 2018. The data sources are listed as follows:

1) ***Discharge Abstract Database (DAD):*** Chronic conditions, or comorbidities (28) came from DAD databases. Also, AMH diagnoses, previous hospitalizations, previous hospitalizations to a designated psychiatric facility, previous hospitalizations to a non-psychiatric facility, homelessness, and police interaction were acquired from the DAD database (29).
2) *National Ambulatory Care Reporting System (NACRS):* Outpatient data, including AMH diagnoses, previous ambulatory visits, elective and nonelective mental health-related emergency department visits (elective visits can be planned in advance or postponed if needed, while nonelective was performed immediately because of an urgent or life-threatening medical condition), history of falling, homelessness and police interaction were accessed from NACRS (30,31).
3) *Practitioner Claims (claims):* AMH diagnoses, physical diagnoses, homelessness, police interaction, and previous physician visits (including family general practitioner, Internal medicine physicians, neurologists, psychiatrists, and others) were accessed from claims.
4) *Pharmaceutical Information Network (PIN):* pharmaceutical and drug dispensations records for mental and addiction diagnoses (only cognitive diagnosis) were collected from PIN.
5) *Alberta Vital Statistics (AVS):* The all-cause mortality data was gained from the AVS database (32).
6) *Registry database:* Insurance status in Alberta obtained from the Registry database.

The data were linked using scrambled versions of the health insurance number. Patient identifiers, names, date of birth, and postal codes were all removed by the AHS. All the codes used in this study were included in supplementary Table 2.

**Cohort Creation**

The dataset consisted of a total of 240,219 unique individuals diagnosed with AMH during the study period in Calgary, Alberta. Most of the studies in epidemiology applied observation and prediction windows with a fixed length to tackle the temporal characteristics. We created a Primary Cohort with a 5-year fixed observation window and a 2-year fixed follow-up (outcome) window (illustrated in Figure 1 (A)). Thus, we included all individuals in Calgary who were diagnosed with AMH and were alive between April 1, 2013, and March 31, 2018, and followed them from March 31, 2018, to March 31, 2020. Like previous studies, a fixed method (Primary Cohort) as standard epidemiology method will be used to identify factors associated with the first occurrence of homelessness and police interaction.

Figure 1(A) illustrates the fixed window length method results in information loss, where patients 2 and 4 are excluded due to first-time outcomes occurring before the index date. It is important to note that even though patient 2 had another outcome within the prediction window, it is not considered as it is not the first occurrence of the outcome. For this method, although the observation window is fixed, individuals whose records start halfway through that fixed window will have observation windows starting from their first record date. As depicted in Figure 1, the initial occurrence of homelessness or police interaction was considered a positive label.

To overcome this limitation, a Secondary Cohort was created using a flexible observation approach (Figure 1 (B)). This method includes all available data for each person up until the day of the outcome of interest, with a minimum history record of either 0, 30, 60, 90, 180, 360, or 720 days (about 2 years). Minimum history record refers to the minimum amount of data (threshold) that must be available for everyone for them to be included in the cohort. In part (B) the subsequent records after the initial outcomes were eliminated using the flexible method. This figure highlights that Patients 2 and 4 were retained in the Secondary Cohort, whereas according to part (A) these two patients were removed using the fixed method.

The observation window started on the first date of an individual's record and extended until the occurrence of the first outcome (positive) or the last entry of the last record for that individual (negative). For individuals who had their first occurrence of the adverse outcome during the observation window, any records beyond that point were excluded from the analysis. This ensured that only relevant information leading up to the first occurrence of the outcome was considered. Consequently, the length of the observation window varied among different individuals.

**Pre-processing**

Each entry of the administrative data is a medical event registered for a patient at a certain time (33), leading to a time series of events that can be physical or mental diagnosis codes (identified as International Statistical Classification of Diseases and Related Health Problems (ICD)), history visits (like clinical visits, emergency department (ED) visits, or hospital visits), or medication codes. Administrative data have static features as well that to not have a time, such as age or sex. The attributes of each record can be categorical or numerical. Each patient may have hundreds of sequential historical events.

In this study, we recorded the presence of various features, such as substance use disorder, within the observation window for each individual. For each event, we assigned a corresponding value representing the number of occurrences. A feature that can have multiple values was called a "multivariable feature" (e.g., number of visits to a psychiatrist). To simplify analysis, we converted the multivariable feature into a "dichotomous feature" by creating two categories based on a threshold (e.g., 0 visits or 1+ visits to a psychiatrist). The outcome variable in this study was dichotomous, where the occurrence of the outcome (e.g., first homelessness or police interaction) was coded as "positive", and the absence of the outcome was coded as "negative."

The adverse outcomes, like homelessness, and police interaction are recorded at a certain time for each patient. The analysis only considered the initial occurrence of a homelessness record and police interaction record for each person, specifically the first record with the relevant ICD code listed in supplementary Table 2. ICD-9 and ICD-10 codes serve as diagnostic coding systems in healthcare to classify and categorize various medical conditions. We utilized these codes to extract data on homelessness (ICD-9 = {V600, V601}, ICD-10= {Z590, Z591}) and police interactions (ICD-9 = {e97.X, v62.5, z65.X, E970-E976, E978, V625}, ICD-10= {Y350-Y357, Z650-Z653}).

**Variables of interest**

1) *Primary Outcome:* The primary outcomes in our study were the first occurrence of homelessness and police interaction.
2) *Covariates:* We considered clinically meaningful factors selected a priori, including demographic characteristics (age and sex), comorbidities (based on the Elixhauser index), and AMH diagnoses. Also, health service utilization, including physician's visits, previous hospital and emergency department visits (with the concern of mental or non-mental health problems) were all included.

**Descriptive Analysis**

Descriptive statistics were employed to summarize the characteristics of the Primary Cohort (as it was created based on the standard epidemiology method) using the dichotomous features. For categorical variables, we reported frequencies and percentages, while for normally distributed continuous variables, we reported means with standard deviations (SD), and for skewed continuous variables, we reported medians with interquartile ranges (IQR).

For the Primary Cohort, since we were interested in identifying the first instance of homelessness and police interaction during this period, individuals who experienced either of these outcomes before March 31, 2018 (or, during the observation window) were excluded from the analysis. After excluding individuals with the outcome of homelessness in the observation window, 2,617 (1.1%) individuals were removed and the remaining records in the Primary Cohort for homelessness (Primary Cohort initial homelessness) consisted of 237,602 individuals. Of those, 1,800 (0.8%) individuals experienced the first occurrence of homelessness. After removing the individuals who had a police interaction outcome within the observation window, 237,141 records remained (3,078 were removed, 1.3%) in the Primary Cohort for the police interaction analysis (Primary Cohort with police-interaction outcome). Only 759 (0.32 %) individuals encountered the police for the first time.

For the Secondary Cohort, we chose a 360-day minimal observation window for our analysis. After applying the minimal 360-day observation window and excluding individuals with less than 360 days of data in the Secondary Cohort, 3,873 individuals (1.6 of all records) were removed. This left a total of 236,346 individuals, of which, 3,814 (1.6%) with the first occurrence of homelessness outcome (about twice as many outcomes as in Primary Cohort) for the analysis (Secondary Cohort with initial homelessness outcome). The same process was applied to the police interaction outcome, resulting in 3,721 individuals (1.5% of all records) being removed and 236,498 records remaining of which, 3,401 (1.4%) individuals with the first police interaction outcome (4.4 times more outcomes than in Primary Cohort) for analysis in cohort 2 (Secondary Cohort with initial police interaction outcome).

**Predictive Factors Analysis**

We employed univariate LR to estimate odds ratios (ORs) with a 95% confidence interval to identify predictive factors for the first occurrence of homelessness and police interaction in Primary Cohort (with a 5-year observation window and 2-year follow-up window) (34). We then applied multivariable LR to examine the joint association of all risk factors for cohort 1. The forward selection method was used to add the variables to the predictive models (35). To validate the predictive models, we randomly split the data into a training set (90% of the sample data) and a test set (10% of the sample) randomly.

**Predictive Model Development**

Three different LR models were trained to predict the occurrence of homelessness and police interaction in the Primary Secondary Cohorts. Model 1 used a 0.5 threshold without any additional techniques, while model 2 used the Youden index (J) to calculate the optimal cut-off point for the LR model. Both model 1 and model 2 were trained using dichotomous features (0, +1) in the observation window. Model 3 used multinomial features and applied PT for normalization. Also, model 3 used ROTE techniques separately for the training set (36,37) and the test set was kept untouched for the final performance report. The Youden index was also used to calculate the optimal cut-off point for model 3.

Along with models 1-3 (LR models with some pre-processing done), ML models, including RF, and XGBoost were evaluated for predicting the first incidence of homelessness and police interaction on both Primary Secondary Cohorts (22,23,27). An alpha level of 0.05 (two-tailed) was used for statistical significance, and a p-value less than 0.05 indicates significant results unlikely due to chance. We submitted several jobs in parallel as array jobs (38). Each job had a unique input to do hyper-parameter optimization. In this experiment, 56,160 trials were performed to determine the optimal hyper-parameters for the ML models. The best results received by these hyper-parameters were illustrated in supplementary Table 1. These models were examined using multivariable features, normalized data, and balanced classes. Data analysis and model development were performed by using Python programming language (version 3.9).

## Results

**Predictors of Homelessness among People with AMH**

In the cohort of 237,602 individuals, 1,800 (0.8%) experienced their first instance of homelessness. Among those, 63.8% (1,149) were men, with a median age of 36.5 years (interquartile range: 21 years). Notably, 78.1% (1,406) had a diagnosis of substance use, 71.1% (1,280) had unplanned mental health emergency department visits, and 64.4% (1,160) visited a psychiatrist. The most prevalent comorbidities were depression, with 74.1% (1,333), and a history of drug abuse, with 60.8% (1,094). (Table 1-a).

In our multivariate analysis of individuals within two years of their first occurrence of homelessness, the following associations were observed (Table 2-a):

1) In terms of demographic characteristics, we found that men had AORs of 1.51 [95% CI: 1.35, 1.69] ($p < 0.01$) compared to women. Individuals aged 60 and above had an odds ratio of 0.73 [95% CI: 0.57, 0.94] ($p = 0.02$), indicating a lower risk compared to the reference group (30-39 years old).
2) Regarding the AMH, it was found that substance use disorder exhibited the highest AORs of 3.70 [95% CI: 3.09, 4.43] ($p < 0.01$), indicating a strong association with the outcome.
3) In terms of health service utilization, we identified that having non-mental health emergency department visits (AORs = 4.21 [95% CI: 3.02, 5.86], $p < 0.01$) and unplanned mental health visits (AORs = 3.33 [95% CI: 2.89, 3.83], $p < 0.01$) were associated with higher odds of the first occurrence of homelessness. Psychiatrist visits had AORs of 1.44 [95% CI: 1.25, 1.67] ($p < 0.01$).
4) Finally, as for comorbidities, our results indicated that the AORs of drug abuse was 2.67 [95% CI: 2.32, 3.09] ($p < 0.01$), indicating a significant association with the outcome.

**Predictors of Police Interactions among people with AMH**

In the analysis, 0.32% (759) of the 237,141 individuals reported a police interaction. Among them, 71.5% (543) were men, with a median age of 36.5 years (interquartile range: 19 years). Anxiety disorder was present in 73.8% (560) and substance use disorder in 66.7% (506). In terms of mental health visits, 61.8% (469) had unplanned MH visits to the ED, 64.0% (486) visited psychiatrists. Also, 78.1% (593) were diagnosed with depression and a history of drug abuse was reported by 49.4% (375).

In our multivariate analysis conducted on individuals who experienced their first instance of homelessness within a two-year period, we identified the following associations (Table 2-b):

1) In relation to demographic characteristics, data showed that males have higher odds of the outcome than females (AOR=2.52, [95% CI: 2.1, 3.02], p<0.01), and older age categories have lower odds compared to the 18-9 and 30-39 age groups, with the 60+ age group having the lowest odds (AOR=0.52, [95% CI: 0.35, 0.78], p<0.01).
2) Concerning AMH features, it was observed that there was a significant association between the police interaction and substance use disorder (AOR=2.83, [95% CI:2.18, 3.69], p<0.01).
3) With respect to health service utilization, we found that the odds of the outcome were significantly higher for ED non-MH visits (AOR=2.68, [95% CI:1.85, 3.88], p<0.01) and unplanned MH visits (AOR=2.06, [95% CI:1.66, 2.54], p<0.01). The AOR for psychiatrist visits was 1.49 [95% CI: 0.19, 1.85] with a statistically significant association (p-value was < 0.01).
4) In terms of comorbidities, it was found that the adjusted OR for valvular disease was 1.97 [95% CI: 1.01, 3.85]. The p-value was 0.05, which was marginally significant, suggesting that the association may not be due to chance. The AOR for drug abuse was 1.83 [95% CI: 1.46, 2.29] with a strong association (p-value of < 0.01).

**Model performance**

**Optimal Minimum Threshold Selection for the Secondary Cohort**

Table 4 for homelessness reveals the following information: for a minimum history record of 0, there are 4,417 individuals with outcomes. For a record of 30, there are 4,266 individuals with outcomes (151 removed). Moving to a record of 60, the count becomes 4,209 (208 removed), and for 90, it is 4,152 (265 removed). A record of 180 shows 4,034 individuals (383 removed), while for 360, the count decreases to 3,814 (603 removed). Lastly, with a minimum history record of 720, the count significantly drops to 3,225, indicating the removal of 1,192 individuals.

Moreover, Table 4 provides the following information regarding police interaction: for a minimum history record of 0, there are 3,837 individuals with outcomes. With a record of 30, the count stands at 3,720 individuals with outcomes (117 removed). When considering a record of 60, the count reduces to 3,687 individuals (150 removed), and for 90, it becomes 3,647 individuals (190 removed). A record of 180 exhibits 3,567 individuals with outcomes (270 removed), while for 360, the count decreases further to 3,401 individuals (436 removed). Finally, with a minimum history record of 720, the count notably drops to 2,894, indicating the removal of 943 individuals.

As shown in Table 4, the performance of the LR model (using dichotomous features) using different minimum record sizes (0, 30, 60, 90, 180, 360, 720) indicates that a threshold of 360 days achieved the highest performance in terms of AUC and sensitivity for homelessness, while shorter and longer window sizes resulted in poorer performance. The flexible window approach considers a person's complete data up to the outcome or the end of their record. However, a short observation window includes individuals with shorter records, potentially causing confusion for the model. Increasing the window threshold consistently reduces the cohort size, which could result in smaller cohorts with large window thresholds, excluding a significant number of positive examples like the fixed window method.

To ensure comparability between the outcomes, a minimum threshold of 360 days was chosen for both homelessness and police interaction to have the same observation window length. This decision was made because the LR model achieved the highest performance using a 180-day minimum threshold for police interaction, while a 360-day minimum threshold ranked third. By using a consistent minimum threshold, the results can be compared effectively between the two outcomes.

**Calibration and Pre-processing Methods**

Upon evaluating the standard LR model, we identified calibration as the pitfall in predictive analytics (39). To address this, we employed the Youden index (J) approach (40,41), which calculates the optimal cut-off point by measuring the difference between the true positive rate and the false positive rate across all potential values (42,43). After implementing the Youden index approach and comparing model 1 and model 2, we observed an increase in sensitivity and AUC scores for the LR model (Table 3 and Figure 2).

To improve the prediction performance, pre-processing methods such as random over-sampling technique (ROTE) to balance classes and power transformation (PT) to normalize data were used in conjunction with these models (44–47).A survey investigation found that model performance was increasingly affected by class imbalances as the degree of class imbalance in the data increased (44), which aligned with our findings.

**Performance of Models in Primary and Secondary Cohorts**

Model 3, which utilized an LR model trained with multinomial features, PT technique (48), ROTE methods, and Youden index, outperformed Model 2 (LR model trained with dichotomous features without using ROTE techniques) across all cohorts. Model 3 demonstrated higher sensitivity (3% for Primary Cohort with initial homelessness outcome, 10% for Primary Cohort with initial police interaction outcome, 1% for Secondary Cohort with initial homelessness outcome, and 3% for Secondary Cohort with initial police interaction outcome) and AUC (1% for Primary Cohort with initial homelessness outcome and Secondary Cohort with initial police interaction outcome) compared to Model 2.

When comparing Primary Cohort and Secondary Cohort, all LR models, especially Model 3, exhibited superior performance (illustrated in Figure 2 and Table 3) in terms of sensitivity, AUC, and precision for Secondary Cohort, which had twice as many outcomes for homelessness and four times as many outcomes for police interaction compared to Primary Cohort. Although both RF and XGBoost models showed potential improvements for Secondary Cohort, they were unable to surpass the performance of LR models in Primary Cohort. Although the ML models did not show improvement for Primary Cohort, there was a significant improvement observed when using ML models for Secondary Cohort. This finding is consistent with previous studies (24), which also reported no improvement using ML models and the standard epidemiology cohort creation method.

We compared two ML models, XGBoost (49) RF (22), with LR models for both cohorts: initial homelessness and initial police interaction. Our findings showed that XGBoost performed better in Secondary Cohort, with a sensitivity of 91% and AUC of 90% for initial homelessness, and a sensitivity of 90% and AUC of 89% for initial police interaction, compared to Model 3 and the RF model.

**Discussion**

We identified several features that are associated with the initial onset of homelessness among Calgary residents diagnosed with AM. Combining these covariates (stated in the variables of interest) achieved superior estimation in this population than if they were considered individually. Our findings were aligned with previous research as demographic characteristics (50–52), AMH features (51,53–56), health service utilization (56,57), and comorbidities, including drug abuse (53,58), HIV/AIDS (51), and alcohol abuse (58) have been previously recognized as risk factors for homelessness in patients diagnosed with AMH. Also, we found several features that were associated with the first occurrence of police interaction for Calgary residents with AMH diagnoses. Our findings were aligned with previous research as demographic characteristics (59–63), AMH features (especially substance use disorder and psychotic disorder) (60,63–66), health service utilization (59–61,64), and general comorbidities (like drug abuse) (65), have been already recognized as risk factors for police interaction in patients diagnosed with AMH. M. Lebenbaum et al. (59) showed the association between overall prior outpatient MH services and police interaction, our study also examined this relationship.

Novel findings emerged from the epidemiological portion of the study:

1) First, we found that individuals with prior visits to a psychiatrist had higher odds of experiencing their first police interaction.
2) Second, it was found that fluid and electrolyte disorders (which were prevalent among patients with chronic alcohol use disorder (67)) were correlated with the first occurrence of homelessness.
3) Third, we identified that valvular disease, fluid and electrolyte disorders, and hearing impairment were significantly associated with the first occurrence of police interaction.
4) Finally, it was observed that non-MH ED visits and non-MH hospitalizations were linked to the initial occurrence of police interaction.

Retaining more adverse outcome information resulted in better results for Secondary Cohort compared to Primary Cohort using both LR and ML models in terms of sensitivity and AUC which was a novel finding. We discovered the importance of calibration (using Youden index) for predictive modeling that was aligned with previous studies (68). We also found using normalization (power transformation) (69) and balancing classes (over-sampling techniques) (70), improved the sensitivity that was aligned with previous studies (71–75). According to these studies, the large degree of the imbalanced classes (in our dataset, the rate was highly imbalanced) lowered the sensitivity, and therefore strategies to improve the model should be considered (72). This study found that the XGBoost model achieved the highest performance in terms of sensitivity and AUC for Secondary Cohort, which was consistent with previous studies (76). However, none of the ML models demonstrated improved performance for the cohort created using the standard epidemiology fixed window method, which aligns with previous studies (24).

We had several limitations that should be acknowledged. First, the pre-processing and linking of the administrative data was challenging due to the volume of data involved. Second, administrative data may contain missing or inaccurate information due to errors in data entry or incomplete reporting. Also, one of the limitations of relying solely on ICD codes to identify individuals with mental health challenges who experience their initial homelessness or police interaction is that they are likely an underestimation of the actual prevalence of this population. Third, administrative data may be subject to selection bias, as it only includes data from individuals who have interacted with the administrative system. This can lead to the underrepresentation of certain groups or the overrepresentation of others. In expressions of ML algorithms, the interplay between overfitting and underfitting was a challenge as the success of these algorithms depends on the selection of the hyper parameters according to the number of observations and features (77). Predictive models can perpetuate or even amplify existing biases in the data used to train them, leading to unfair or discriminatory outcomes.

In conclusion, we discovered that visits to a psychiatrist, non-MH ED, and non-MH hospitalization were linked to the first police interaction. Additionally, we identified three previously unexplored comorbidities (valvular disease, fluid and electrolyte disorders, and hearing impairment) associated with the initial police interaction and one new comorbidity (fluid and electrolyte disorders) associated with the onset of homelessness.

Furthermore, our findings indicated that implementing a flexible window for cohort creation and including a greater amount of adverse outcome information resulted in improved performance in terms of sensitivity, AUC and precision. Additionally, using the flexible window cohort creation method, incorporating multivariable features, and utilizing normalization, ROTE, and XGBoost classification model resulted in superior prediction performance compared to other methods. These findings have the potential to improve treatment decisions in healthcare. Further research was needed to explore the technical and clinical applications of these cohort creation methods to improve outcomes.

## Conflict of Interest

The authors declare that the research was conducted in the absence of any commercial or financial relationships that could be construed as a potential conflict of interest.

## Author Contributions

**F.SH.:** Conceptualization, Methodology, Software, Validation, Formal analysis, Investigation, Resources, Data Curation, Writing - Original Draft, Writing - Review & Editing, Visualization.

**M.E.M.:** Conceptualization, Methodology, Writing - Review and Editing, Supervision, Project administration.

**G.M.:** Conceptualization, Methodology, Validation, Formal analysis, Investigation, Resources, Data Curation, Writing - Original Draft, Writing - Review & Editing, Visualization, Supervision, Project administration, Funding acquisition

**D.S.:** Conceptualization, Methodology, Data Curation, Writing, - Review and Editing, Supervision, Project administration, Funding acquisition.

## Funding

This study received funding from the Calgary Health Foundation Institute.## Data Availability Statement

The data was provided by AHS. Due to the sensitivity of the information, AHS has restricted the use of the data and they are not publicly available. To obtain the data, researchers will need approval from a certified research ethics board in Alberta as well as a Data Sharing Agreement from AHS through ethics boards. The websites for ethics boards and data sharing agreements are respectively located at https://hreba.ca/ and https://www.albertahealthservices.ca/research/Page16074.aspx.

To include, in this order: Accession codes (https://github.com/Fuzzy-sh/Machine-Learning-Risk-Estimation-and-Prediction-of-Homelessness-and-Police-interaction.git)

# References

1. Su C, Xu Z, Pathak J, Wang F. Deep learning in mental health outcome research: a scoping review. Translational Psychiatry (2020) 10:1-116 p. doi: 10.1038/s41398-020-0780-3
2. McIntosh AM, Stewart R, John A, Smith DJ, Davis K, Sudlow C, et al. Data science for mental health: a UK perspective on a global challenge. The Lancet Psychiatry (2016) 3:10. p. 993–8. doi: 10.1016/S2215-0366(16)30089-X.
3. These innovative projects are tackling homelessness around the world. World Economic Forum (2021). https://www.weforum.org/agenda/2021/10/innovative-projects-tackling-homelessness-around-the-world/ [Accessed Jun 15, 2023].
4. Laniyonu A, Goff PA. Measuring disparities in police use of force and injury among persons with serious mental illness. BMC Psychiatry (2021) 21:1-500 p. doi: 10.1186/s12888-021-03510-w
5. Canada PHA of. Mental Illness in Canada - Infographic (2020). https://www.canada.ca/en/public-health/services/publications/diseases-conditions/mental-illness-canada-infographic.html [Accessed Nov 22, 2022].
6. Gravel R, Béland Y. The Canadian Community Health Survey: mental health and well-being. The Canadian Journal of Psychiatry (2005) 50:10. p. 573–9.
7. Government of Canada SC. Characterizing people experiencing homelessness and trends in homelessness using population-level emergency department visit data in Ontario, Canada (2021). https://www150.statcan.gc.ca/n1/pub/82-003-x/2021001/article/00002-eng.htm [Accessed Jan 4, 2023].
8. Draine J, Salzer MS, Culhane DP, Hadley TR. Role of Social Disadvantage in Crime, Joblessness, and Homelessness Among Persons With Serious Mental Illness. Psychiatric Services (2002) 53:5. p. 565-73. doi: 10.1176/appi.ps.53.5.565
9. 2020: Likely a Record Breaking Year for Deaths by Police in Canada. B. Law (2020). https://criminallawoshawa.com/2020-likely-a-record-breaking-year-for-deaths-by-police-in-canada/ [Accessed Jan 4, 2023].
10. Canadian Mental Health Association. (2005). Police and mental illness: Increased interactions. British Columbia: CMHA BC Division. https://www.publicsafety.gc.ca/lbrr/archives/cnmcs-plcng/cn34078-2005-1-eng.pdf.
11. Hoch JS, Hartford K, Heslop L, Stitt L. Mental Illness and Police Interactions in a Mid-Sized Canadian City: What the Data Do and Do Not Say. Canadian Journal of Community Mental Health (2009) 28:1. p. 49–66. doi: 10.7870/cjcmh-2009-0005
12. Wallace D, Wang X. Does in-prison physical and mental health impact recidivism? SSM - Population Health (2020) 11. 100569 p. doi: 10.1016/j.ssmph.2020.100569
13. Yoo R, Krawczyk N, Johns E, McCormack RP, Rotrosen J, Mijanovich T, et al. Association of substance use characteristics and future homelessness among emergency department patients with drug use or unhealthy alcohol use: Results from a linked data longitudinal cohort analysis. Substance Abuse (2022) 43:1. p. 1100–9. doi: 10.1080/08897077.2022.2060445
14. Von Wachter T, Bertrand M, Pollack H, Rountree J, Blackwell B. Predicting and preventing homelessness in Los Angeles. California Policy Lab and University of Chicago Poverty Lab (2019). https://www.capolicylab.org/wp-content/uploads/2019/12/Predicting_and_Preventing_Homelessness_in_Los_Angeles.pdf
15. Wang F, Lee N, Hu J, Sun J, Ebadollahi S. Towards heterogeneous temporal clinical event pattern discovery: a convolutional approach. In: 18th ACM SIGKDD international conference on Knowledge discovery and data mining (2012). p. 453-461. doi: 10.1145/2339530.2339605
16. Wang Y, Ng K, Byrd RJ, Hu J, Ebadollahi S, Daar Z, et al. Early detection of heart failure with varying prediction windows by structured and unstructured data in electronic health records. In:

2015 37th Annual International Conference of the IEEE Engineering in Medicine and Biology Society (EMBC) (2015). p. 2530–3. doi: 10.1109/EMBC.2015.7318907
17. Ng K, Ghoting A, Steinhubl SR, Stewart WF, Malin B, Sun J. PARAMO: A PARAllel predictive MOdeling platform for healthcare analytic research using electronic health records. Journal of Biomedical Informatics (2014) 48. p. 160–70. doi: 10.1016/j.jbi.2013.12.012
18. Ng K, Steinhubl SR, DeFilippi C, Dey S, Stewart WF. Early detection of heart failure using electronic health records: practical implications for time before diagnosis, data diversity, data quantity, and data density. Circulation: Cardiovascular Quality and Outcomes (2016) 9:6. p. 649-58. doi: 10.1161/CIRCOUTCOMES.116.002797
19. Chen R, Stewart WF, Sun J, Ng K, Yan X. Recurrent neural networks for early detection of heart failure from longitudinal electronic health record data: implications for temporal modeling with respect to time before diagnosis, data density, data quantity, and data type. Circulation: Cardiovascular Quality and Outcomes (2019) 12:10. e005114 p. doi: 10.1161/CIRCOUTCOMES.118.005114
20. Gupta M, Poulain R, Phan TL, Bunnell HT, Beheshti R. Flexible-Window Predictions on Electronic Health Records. In: AAAI Conference on Artificial Intelligence (2022) 36:11. p. 12510-12516. doi: 10.1609/aaai.v36i11.21520
21. Ranganathan P, Pramesh CS, Aggarwal R. Common pitfalls in statistical analysis: Logistic regression. Perspectives in clinical research (2017) 8:3-148 p. doi: 10.4103/picr.PICR_87_17
22. King C, Strumpf E. Applying random forest in a health administrative data context: a conceptual guide. Health Services and Outcomes Research Methodology (2022) 22:1. p. 96–117. doi: 10.1007/s10742-021-00255-7
23. Tiwari P, Colborn KL, Smith DE, Xing F, Ghosh D, Rosenberg MA. Assessment of a Machine Learning Model Applied to Harmonized Electronic Health Record Data for the Prediction of Incident Atrial Fibrillation. JAMA Network Open (2020) 3:1.e1919396 p. doi: 10.1001/jamanetworkopen.2019.19396
24. Choi SB, Lee W, Yoon JH, Won JU, Kim DW. Ten-year prediction of suicide death using Cox regression and machine learning in a nationwide retrospective cohort study in South Korea. Journal of affective disorders (2018) 231. p. 8-14. doi: 10.1016/j.jad.2018.01.019
25. Breiman L. Random Forests. Machine Learning (2001) 45:1. p. 5–32. doi: 10.1023/A:1010933404324
26. Ooka T, Johno H, Nakamoto K, Yoda Y, Yokomichi H, Yamagata Z. Random forest approach for determining risk prediction and predictive factors of type 2 diabetes: large-scale health check-up data in Japan. BMJ Nutrition, Prevention & Health (2021) 4:1. 140 p. doi: 10.1136/bmjnph-2020-000200
27. Ogunleye A, Wang QG. XGBoost Model for Chronic Kidney Disease Diagnosis. IEEE/ACM Transactions on Computational Biology and Bioinformatics (2019) 17:6. p. 2131–40. doi: 10.1109/TCBB.2019.2911071
28. van Walraven C, Austin PC, Jennings A, Quan H, Forster AJ. A Modification of the Elixhauser Comorbidity Measures Into a Point System for Hospital Death Using Administrative Data. Medical Care (2009) 47:6. p. 626-33.
29. Discharge Abstract Database metadata (DAD) | CIHI. https://www.cihi.ca/en/discharge-abstract-database-metadata-dad [Accessed Jan 31, 2022].
30. NACRS Data Elements, 2021–2022. https://www.cihi.ca/sites/default/files/rot/nacrs-data-elements-2021-2022-en.pdf [Accessed Jan 31, 2022].
31. National Ambulatory Care Reporting System metadata (NACRS) | CIHI. https://www.cihi.ca/en/national-ambulatory-care-reporting-system-metadata-nacrs [Accessed Jan 31, 2022].
32. Vital statistics form. https://www.alberta.ca/vital-statistics-forms.aspx [Accessed Feb 16, 2022].

33. Liu L, Li H, Hu Z, Shi H, Wang Z, Tang J, Zhang M. Learning hierarchical representations of electronic health records for clinical outcome prediction. In: AMIA Annual Symposium Proceedings (2019).597 p. American Medical Informatics Association.
34. Szumilas M. Explaining odds ratios. Journal of the Canadian academy of child and adolescent psychiatry (2010) 19:3. p. 227-9.
35. Chowdhury MZI, Turin TC. Variable selection strategies and its importance in clinical prediction modelling. Family medicine and community health (2020) 8:1. e000262 p. doi: 10.1136/fmch-2019-000262.
36. Alsinglawi B, Alshari O, Alorjani M, Mubin O, Alnajjar F, Novoa M, et al. An explainable machine learning framework for lung cancer hospital length of stay prediction. Scientific reports (2022) 12:1. 607 p. doi: 10.1038/s41598-021-04608-7
37. Garcia-Carretero R, Roncal-Gomez J, Rodriguez-Manzano P, Vazquez-Gomez O. Identification and Predictive Value of Risk Factors for Mortality Due to Listeria monocytogenes Infection: Use of Machine Learning with a Nationwide Administrative Data Set. Bacteria (2022) 1:1. p. 12–32. doi: 10.3390/bacteria1010003
38. ARC Cluster Guide - RCSWiki. https://rcs.ucalgary.ca/ARC_Cluster_Guide [Accessed Jan 30, 2023].
39. Van Calster B, McLernon DJ, van Smeden M, Wynants L, Steyerberg EW, Bossuyt P, et al. Calibration: the Achilles heel of predictive analytics. BMC Medicine (2019) 17:1. 230 p. doi: https://doi.org/10.1186/s12916-019-1466-7
40. Unal I. Defining an Optimal Cut-Point Value in ROC Analysis: An Alternative Approach. Computational and mathematical methods in medicine (2017) 2017. 3762651 p. doi: 10.1155/2017/3762651
41. Youden WJ. Index for rating diagnostic tests. Cancer (1950) 3:1. p. 32–5. Doi: 10.1002/1097-0142(1950)3:1<32::AID-CNCR2820030106>3.0.CO;2-3
42. Perkins NJ, Schisterman EF. The Youden Index and the optimal cut-point corrected for measurement error. Biometrical Journal: Journal of Mathematical Methods in Biosciences (2005) 47:4. p. 428-41. doi: 10.1002/bimj.200410133
43. Fluss R, Faraggi D, Reiser B. Estimation of the Youden Index and its associated cutoff point. Biometrical Journal: Journal of Mathematical Methods in Biosciences (2005) 47:4. p. 458-72. doi: 10.1002/bimj.200410135
44. Japkowicz N, Stephen S. The class imbalance problem: A systematic study. Intelligent Data Analysis (2002) 6:5. p. 429-49.
45. Krittanawong C, Virk HUH, Kumar A, Aydar M, Wang Z, Stewart MP, et al. Machine learning and deep learning to predict mortality in patients with spontaneous coronary artery dissection. Scientific reports (2021) 11:1. 8992 p.
46. Mahmoudi E, Kamdar N, Kim N, Gonzales G, Singh K, Waljee AK. Use of electronic medical records in development and validation of risk prediction models of hospital readmission: systematic review. BMJ (2020) 369. Doi: 10.1136/bmj.m958
47. Singh D, Singh B. Investigating the impact of data normalization on classification performance. Applied Soft Computing (2020) 97. 105524 p. doi: 10.1016/j.asoc.2019.105524
48. Weisberg S. Yeo-Johnson power transformations. Department of Applied Statistics, University of Minnesota (2001). 2003 p.
49. Chen T, Guestrin C. Xgboost: A scalable tree boosting system. In: 22nd acm sigkdd international conference on knowledge discovery and data mining (2016). p. 785-94. doi: 10.1145/2939672.2939785
50. Marpsat M. An Advantage with Limits: The Lower Risk for Women of Becoming Homeless. Population: An English Selection (2000). p. 247–91.

51. Culhane DP, Gollub E, Kuhn R, Shpaner M. The co-occurrence of AIDS and homelessness: results from the integration of administrative databases for AIDS surveillance and public shelter utilisation in Philadelphia. Journal of Epidemiology & Community Health (2001). 55:7. p. 515–20. doi: 10.1136/jech.55.7.515
52. Caton CLM, Dominguez B, Schanzer B, Hasin DS, Shrout PE, Felix A, et al. Risk factors for long-term homelessness: findings from a longitudinal study of first-time homeless single adults. American journal of public health (2005) 95:10. p.1753–9.
53. Johnson TP, Freels SA, Parsons JA, Vangeest JB. Substance abuse and homelessness: social selection or social adaptation? Addiction (1997) 92:4. p. 437–45. doi: 10.1111/j.1360-0443.1997.tb03375.x
54. Schütz CG. Homelessness and Addiction: Causes, Consequences and Interventions. Current Treatment Options in Psychiatry (2016) 3. p. 306-13. doi: 10.1007/s40501-016-0090-9
55. Ayano G, Shumet S, Tesfaw G, Tsegay L. A systematic review and meta-analysis of the prevalence of bipolar disorder among homeless people. BMC Public Health (2020). 20:1. 731 p. doi: 10.1186/s12889-020-08819-x
56. Folsom DP, Hawthorne W, Lindamer L, Gilmer T, Bailey A, Golshan S, et al. Prevalence and Risk Factors for Homelessness and Utilization of Mental Health Services Among 10,340 Patients With Serious Mental Illness in a Large Public Mental Health System. American Journal of Psychiatry (2005) 162:2. p. 370–6. doi: 10.1176/appi.ajp.162.2.370
57. Yue D, Pourat N, Essien EA, Chen X, Zhou W, O'Masta B. Differential associations of homelessness with emergency department visits and hospitalizations by race, ethnicity, and gender. Health Services Research (2022) 57. p. 249–62. doi: 10.1111/1475-6773.14009
58. Shelton KH, Taylor PJ, Bonner A, van den Bree M. Risk Factors for Homelessness: Evidence From a Population-Based Study. Psychiatric services (2009) 60:4. p.465–72. doi: 10.1176/ps.2009.60.4.465
59. Lebenbaum M, Kouyoumdjian F, Huang A, Kurdyak P. The Association Between Prior Mental Health Service Utilization and Risk of Recidivism among Incarcerated Ontario Residents. The Canadian Journal of Psychiatry (2022). 07067437221140385 p. doi: 10.1177/07067437221140385
60. Williams-Butler A, Liu FY, Howell T, Menon SE, Quinn CR. Racialized Gender Differences in Mental Health Service Use, Adverse Childhood Experiences, and Recidivism Among Justice-Involved African American Youth. Race and Social Problems (2022). p.1-4. doi: 10.1007/s12552-022-09360-9
61. Evans E, Huang D, Hser YI. High-Risk Offenders Participating in Court-Supervised Substance Abuse Treatment: Characteristics, Treatment Received, and Factors Associated with Recidivism. The journal of behavioral health services & research (2011) 38:4. p.510–25. doi: 10.1007/s11414-011-9241-3
62. McCoy AM, Como JJ, Greene G, Laskey SL, Claridge JA. A novel prospective approach to evaluate trauma recidivism: The concept of the past trauma history. Journal of Trauma and Acute Care Surgery (2013) 75:1. p. 116-21. doi: 10.1097/TA.0b013e31829231b7
63. Piel JL, Schouten R. Violence Risk Assessment. Mental Health Practice and the Law. New York, NY: Oxford University Press (2017). p. 39-60.
64. Morrissey JP, Cuddeback GS, Cuellar AE, Steadman HJ. The Role of Medicaid Enrollment and Outpatient Service Use in Jail Recidivism Among Persons With Severe Mental Illness. Psychiatric Services (2007) 58:6. p.794–801.
65. Colins O, Vermeiren R, Vahl P, Markus M, Broekaert E, Doreleijers T. Psychiatric Disorder in Detained Male Adolescents as Risk Factor for Serious Recidivism. The Canadian Journal of Psychiatry (2011) 56:1. p. 44–50.

66. McReynolds LS, Schwalbe CS, Wasserman GA. The Contribution of Psychiatric Disorder to Juvenile Recidivism. Criminal Justice and Behavior (2010) 37:2. p. 204–16. doi: 10.1177/0093854809354961
67. Palmer BF, Clegg DJ. Electrolyte Disturbances in Patients with Chronic Alcohol-Use Disorder. N Engl J Med (2017) 377. p. 1368–77. doi: 10.1056/NEJMra1704724
68. Unnikrishnan VK, Choudhari KS, Kulkarni SD, Nayak R, Kartha VB, Santhosh C. Analytical predictive capabilities of Laser Induced Breakdown Spectroscopy (LIBS) with Principal Component Analysis (PCA) for plastic classification. RSC Advances (2013) 3:48. p. 25872–80. doi: 10.1039/C3RA44946G
69. Dairi A, Harrou F, Zeroual A, Hittawe MM, Sun Y. Comparative study of machine learning methods for COVID-19 transmission forecasting. Journal of Biomedical Informatics (2021) 118. 103791 p. doi: 10.1016/j.jbi.2021.103791
70. Mufti HN, Hirsch GM, Abidi SR, Abidi SSR. Exploiting Machine Learning Algorithms and Methods for the Prediction of Agitated Delirium After Cardiac Surgery: Models Development and Validation Study. JMIR medical informatics (2019) 7:4. e14993 p. doi:10.2196/14993
71. Bragg WH. LXXIII. On the absorption of α rays, and on the classification of the α rays from radium. The London, Edinburgh, and Dublin Philosophical Magazine and Journal of Science (1904) 8:48. p. 719–25.
72. Japkowicz N, Stephen S. The class imbalance problem: A systematic study. Intelligent Data Analysis (2002) 6:5. p. 429–49. doi: 10.3233/IDA-2002-6504
73. Mahmoudi E, Kamdar N, Kim N, Gonzales G, Singh K, Waljee AK. Use of electronic medical records in development and validation of risk prediction models of hospital readmission: systematic review. BMJ (2020) 369. m958 p. doi: 10.1136/bmj.m958
74. Garcia-Carretero R, Roncal-Gomez J, Rodriguez-Manzano P, Vazquez-Gomez O. Identification and Predictive Value of Risk Factors for Mortality Due to Listeria monocytogenes Infection: Use of Machine Learning with a Nationwide Administrative Data Set. Bacteria (2022) 1:1. p. 12–32. doi: 10.3390/bacteria1010003
75. Alsinglawi B, Alshari O, Alorjani M, Mubin O, Alnajjar F, Novoa M, et al. An explainable machine learning framework for lung cancer hospital length of stay prediction. Scientific reports. (2022) 12:1. 607 p. doi: 10.1038/s41598-021-04608-7
76. Yang H, Li J, Liu S, Yang X, Liu J. Predicting Risk of Hypoglycemia in Patients With Type 2 Diabetes by Electronic Health Record–Based Machine Learning: Development and Validation. JMIR Medical Informatics (2022) 10:6. e36958 p. doi:10.2196/36958
77. Shameer K, Johnson KW, Glicksberg BS, Dudley JT, Sengupta PP. Machine learning in cardiovascular medicine: are we there yet? Heart (2018) 104:14. p. 1156–64. doi: 10.1136/heartjnl-2017-311198

**Table 1-a    Characteristics of individuals in Calgary, Alberta, Canada who were diagnosed with AMH between April 1, 2013, and March 31, 2018, and the difference between those who experienced homelessness for the first time during the two-year follow-up period and those who did not. Abbreviations: AMH, addiction, and mental health; ED, emergency department; MH, mental health; H, hospital; GP: general practitioner; IM: internal medicine; IQR, interquartile range.**

|  | The first occurrence of the homelessness | | Total=237,602 |
|---|---|---|---|
| Characteristic | Yes= 1,800 (0.8%) | No=235,802 (99.2%) | |
| Gender | | | |
|    Male | 1,149 (63.8) | 100,879 (42.8) | 102,028 (42.9) |
|    Female | 651 (36.2) | 134,923 (57.2) | 135,574 (57.1) |
| Age | [IQR:21] [median:36.5] | [IQR:22] [median:42] | [IQR:22] [median:42] |
| Age category | | | |
|    18-29 | 520 (28.9) | 49,322 (20.9) | 49,842 (21.0) |
|    30-39 | 511 (28.4) | 56,686 (24.0) | 57,197 (24.1) |
|    40-49 | 346 (19.2) | 52,795 (22.4) | 53,141 (22.4) |
|    50-59 | 327 (18.2) | 52,912 (22.4) | 53,239 (22.4) |
|    60+ | 96 (5.3) | 24,087 (10.2) | 24,183 (10.2) |
| AMH features | | | |
|    Substance use disorder | 1,406 (78.1) | 44,742 (19.0) | 46,148 (19.4) |
|    Anxiety disorder | 1,240 (68.9) | 154,201 (65.4) | 155,441 (65.4) |
|    Mood disorder | 1,162 (64.6) | 98,167 (41.6) | 99,329 (41.8) |
|    Other psychiatric disorders | 650 (36.1) | 60,521 (25.7) | 61,171 (25.7) |
|    Psychotic disorder | 388 (21.6) | 8,845 (3.8) | 9,233 (3.9) |
|    Deliberate self-harm | 57 (3.2) | 1,398 (0.6) | 1,455 (0.6) |
|    Cognitive disorders | 19 (1.1) | 1,623 (0.7) | 1,642 (0.7) |
| Health Service Utilization | | | |
| ED | | | |
|    ED Non-MH visits | 1,757 (97.6) | 199,688 (84.7) | 201,445 (84.8) |
|    Unplanned MH visits | 1,280 (71.1) | 35,329 (15.0) | 36,609 (15.4) |
|    ED visits | 725 (40.3) | 42,449 (18.0) | 43,174 (18.2) |
| Hospital | | | |
|    Non-MH hospitalized | 478 (26.6) | 7692 (3.3) | 8,170 (3.4) |
|    MH hospitalized | 432 (24.0) | 8816 (3.7) | 9,248 (3.9) |
|    H visits | 176 (9.8) | 4217 (1.8) | 4,393 (1.8) |
| Clinic history visits | | | |
|    GP | 1,717 (95.4) | 225,162 (95.5) | 226,879 (95.5) |
|    Other physicians | 1,700 (94.4) | 216,253 (91.7) | 217,953 (91.7) |
|    Psychiatrist | 1,160 (64.4) | 60,851 (25.8) | 62,011 (26.1) |
|    IM | 950 (52.8) | 122,734 (52.0) | 123,684 (52.1) |
|    Neurologist | 217 (12.1) | 22,850 (9.7) | 23,067 (9.7) |
|    Pharmacy | 0 (0.0) | 152 (0.1) | 152 (0.1) |
| Comorbidities | | | |
|    Depression | 1,333 (74.1) | 134432 (57.0) | 135,765 (57.1) |
|    Drug abuse | 1,094 (60.8) | 16574 (7.0) | 17,668 (7.4) |
|    Tobacco use | 886 (49.2) | 41250 (17.5) | 42,136 (17.7) |
|    Alcohol abuse | 708 (39.3) | 9076 (3.8) | 9,784 (4.1) |
|    Psychotic disorders | 570 (31.7) | 14442 (6.1) | 15,012 (6.3) |

**Table 1-b** Characteristics of individuals in Calgary, Alberta, Canada who were diagnosed with AMH between April 1, 2013, and March 31, 2018, and the difference between those who experienced police interaction for the first time during the two-year follow-up period and those who did not. Abbreviations: AMH, addiction, and mental health; ED, emergency department; MH, mental health; H, hospital; GP: general practitioner; IM: internal medicine; IQR, interquartile range.

|  | The first occurrence of the police interaction | | Total=237,141 |
|---|---|---|---|
| Characteristic | Yes= 759 (0.32 %) | No=236,382 (99.68%) | |
| Gender | | | |
|     Male | 543 (71.5) | 100958 (42.7) | 101,501 (42.8) |
|     Female | 216 (28.5) | 135424 (57.3) | 135,640 (57.2) |
| Age | [IQR:19] [median:36.5] | [IQR:22] [median:42] | [IQR:22] [median:42] |
| Age category | | | |
|     18-29 | 239 (31.5) | 49258 (20.8) | 49,497 (20.9) |
|     30-39 | 242 (31.9) | 56666 (24.0) | 56,908 (24.0) |
|     40-49 | 141 (18.6) | 52946 (22.4) | 53,087 (22.4) |
|     50-59 | 105 (13.8) | 53275 (22.5) | 53,380 (22.5) |
|     60+ | 32 (4.2) | 24237 (10.3) | 24,269 (10.2) |
| AMH features | | | |
|     Anxiety disorder | 560 (73.8) | 154760 (65.5) | 155,320 (65.5) |
|     Mood disorder | 518 (68.2) | 98588 (41.7) | 99,106 (41.8) |
|     Substance use disorder | 506 (66.7) | 45606 (19.3) | 46,112 (19.4) |
|     Other psychiatric disorders | 296 (39.0) | 60831 (25.7) | 61,127 (25.8) |
|     Psychotic disorder | 172 (22.7) | 8827 (3.7) | 8,999 (3.8) |
|     Deliberate self-harm | 40 (5.3) | 1486 (0.6) | 1,526 (0.6) |
|     Cognitive disorders | 7 (0.9) | 1676 (0.7) | 1,683 (0.7) |
| Health Service Utilization | | | |
| ED | | | |
|     ED Non-MH visits | 726 (95.7) | 200245 (84.7) | 200,947 (84.7) |
|     Unplanned MH visits | 469 (61.8) | 35899 (15.2) | 36,968 (15.6) |
|     ED visits | 334 (44.0) | 42596 (18.0) | 43,187 (18.2) |
| Hospital | | | |
|     Unplanned MH hospitalized | 184 (24.2) | 36499 (15.4) | 9,611 (4.1) |
|     Non-MH hospitalized | 161 (21.2) | 200221 (84.7) | 8,617 (3.6) |
|     H visits | 51 (6.7) | 4522 (1.9) | 4,573 (1.9) |
| Clinic history visits | | | |
|     GP | 719 (94.7) | 225684 (95.5) | 226,403 (95.5) |
|     Other physicians | 710 (93.5) | 216822 (91.7) | 217,532 (91.7) |
|     Psychiatrist | 486 (64.0) | 61014 (25.8) | 61,500 (25.9) |
|     IM | 369 (48.6) | 123397 (52.2) | 123,766 (52.2) |
|     Neurologist | 87 (11.5) | 23049 (9.8) | 23,136 (9.8) |
|     Pharmacy | 1 (0.1) | 150 (0.1) | 152 (0.1) |
| Comorbidities | | | |
|     Depression | 593 (78.1) | 135006 (57.1) | 135,599 (57.2) |
|     Drug abuse | 375 (49.4) | 17238 (7.3) | 17,613 (7.4) |
|     Psychotic disorders | 237 (31.2) | 14320 (6.1) | 14,557 (6.1) |
|     Tobacco use | 332 (43.7) | 41807 (17.7) | 42,139 (17.8) |
|     Alcohol abuse | 244 (32.1) | 9837 (4.2) | 10,081 (4.3) |

**Table 2-a**   Associations between clinical risk factors and the first occurrence of all-cause homelessness during a period of seven years between, April 1, 2013, to March 31, 2020, in Calgary, Alberta diagnosed with AMH who were observed for at least 360 days during this period. Abbreviations: AMH, addiction, and mental health; ED, emergency department; MH, mental health; H, hospital; GP: general practitioner; IM: internal medicine.

|  | Univariate |  | Multivariate |  |
|---|---|---|---|---|
| Characteristic | ORs (95% CI) | P | AORs (95% CI) | P |
| Gender |  |  |  |  |
|     Male | 2.36 (2.14, 2.60) | < 0.01 | 1.51 (1.35, 1.69) | < 0.01 |
|     Female | 1 [Reference] |  | 1 [Reference] |  |
| Age category |  |  |  |  |
|     60+ | 0.44 (0.36, 0.55) | < 0.01 | 0.73 (0.57, 0.94) | 0.02 |
|     50-59 | 0.69 (0.60, 0.79) | < 0.01 | 0.95 (0.81, 1.12) | 0.53 |
|     40-49 | 0.73 (0.63, 0.83) | < 0.01 | 0.96 (0.82, 1.07) | 0.58 |
|     30-39 | 1 [Reference] |  | 1 [Reference] |  |
|     18-29 | 1.17 (1.03, 1.32) | 0.06 | 0.94 (0.82, 1.07) | 0.35 |
| AMH features |  |  |  |  |
|   Substance use disorder | 15.24 (13.62, 17.05) | < 0.01 | 3.70 (3.09, 4.43) | < 0.01 |
|   Mood disorder | 2.55 (2.32, 2.81) | < 0.01 | 1.5 (1.28, 1.76) | < 0.01 |
|   Psychotic disorder | 7.05 (6.29, 7.90) | < 0.01 | 1.41 (1.15, 1.74) | < 0.01 |
|   Other psychiatric disorders | 1.64 (1.49, 1.80) | < 0.01 | 1.03 (0.91, 1.17) | 0.65 |
|   Anxiety disorder | 1.17 (1.06, 1.30) | < 0.01 | 1.02 (0.90, 1.16) | 0.72 |
|   Deliberate self-harm | 5.48 (4.19, 7.18) | < 0.01 | 0.67 (0.49, 0.90) | <0.01 |
|   Cognitive disorders | 1.54 (0.98, 2.43) | 0.06 | 0.60 (0.35, 1.02) | 0.06 |
| Health Service Utilization |  |  |  |  |
|   ED |  |  |  |  |
|     ED Non-MH visits | 7.39 (5.46, 10.00) | < 0.01 | 4.21 (3.02, 5.86) | < 0.01 |
|     Unplanned MH visits | 13.97 (12.61, 15.48) | < 0.01 | 3.33 (2.89, 3.83) | < 0.01 |
|     ED visits | 3.07 (2.79, 3.38) | < 0.01 | 0.83 (0.73, 0.94) | < 0.01 |
|   Hospital |  |  |  |  |
|     Non-MH hospitalized | 10.72 (9.63, 11.93) | < 0.01 | 1.14 (0.98, 1.34) | 0.09 |
|     H visits | 5.95 (5.08, 6.97) | < 0.01 | 0.99 (0.82, 1.20) | 0.90 |
|     MH hospitalized | 8.13 (7.28, 9.08) | < 0.01 | 0.70 (0.60, 0.82) | < 0.01 |
|   Clinic history visits |  |  |  |  |
|     Psychiatrist | 5.21 (4.73, 5.74) | < 0.01 | 1.44 (1.25, 1.67) | < 0.01 |
|     Other physicians | 1.54 (1.26, 1.88) | 0.02 | 1.10 (0.70, 1.73) | 0.67 |
|     IM | 1.03 (0.94, 1.13) | 0.62 | 0.79 (0.71, 0.89) | < 0.01 |
|     Neurologist | 1.28 (1.11, 1.47) | < 0.01 | 0.77 (0.64, 0.93) | < 0.01 |
|     GP | 0.98 (0.78, 1.22) | < 0.01 | 0.76 (0.46, 1.26) | 0.29 |
| Comorbidities |  |  |  |  |
|   Drug abuse | 20.50 (18.62, 22.56) | < 0.01 | 2.67 (2.32, 3.09) | < 0.01 |
|   HIV/AIDS | 8.28 (5.60, 12.25) | < 0.01 | 2.25 (1.39, 3.62) | < 0.01 |
|   Lymphoma | 1.82 (1.03, 3.22) | 0.04 | 2.01 (1.08, 3.73) | 0.03 |
|   Alcohol abuse | 16.20 (14.70, 17.84) | < 0.01 | 1.61 (1.41, 1.84) | < 0.01 |
|   Fluid and electrolyte disorders | 5.78 (5.22, 6.40) | < 0.01 | 1.52 (1.33, 1.73) | < 0.01 |

**Table 2-b** Associations between clinical risk factors and the first occurrence of all-cause police interaction during a period of seven years between, April 1, 2013, to March 31, 2020, in Calgary, Alberta diagnosed with AMH who were observed for at least 360 days during this period. Abbreviations: AMH, addiction, and mental health; ED, emergency department; MH, mental health; H, hospital; GP: general practitioner; IM: internal medicine.

|  | Univariate | | Multivariate | |
|---|---|---|---|---|
| Characteristic | ORs (95% CI) | P | AORs (95% CI) | P |
| Gender | | | | |
|   Male | 3.37 (2.88, 3.95) | < 0.01 | 2.52 (2.1, 3.02) | < 0.01 |
|   Female | 1 [Reference] | | 1 [Reference] | |
| Age category | | | | |
|   18-29 | 1.14 (0.95, 1.36) | 0.16 | 0.89 (0.73, 1.08) | 0.23 |
|   30-39 | 1 [Reference] | | 1 [Reference] | |
|   40-49 | 0.62 (0.51, 0.77) | < 0.01 | 0.75 (0.6, 0.94) | < 0.01 |
|   50-59 | 0.46 (0.37, 0.58) | < 0.01 | 0.59 (0.45, 0.76) | < 0.01 |
|   60+ | 0.31 (0.21, 0.45) | < 0.01 | 0.52 (0.35, 0.78) | < 0.01 |
| AMH features | | | | |
|   Substance use disorder | 8.37 (7.19, 9.73) | < 0.01 | 2.83 (2.18, 3.69) | < 0.01 |
|   Psychotic disorder | 7.55 (6.36, 8.96) | < 0.01 | 1.73 (1.26, 2.36) | < 0.01 |
|   Mood disorder | 3.00 (2.58, 3.50) | < 0.01 | 1.60 (1.26, 2.03) | < 0.01 |
|   Deliberate self-harm | 8.79 (6.37, 12.14) | < 0.01 | 1.29 (0.88, 1.88) | 0.19 |
|   Anxiety disorder | 1.48 (1.26, 1.75) | < 0.01 | 1.28 (1.05, 1.55) | < 0.01 |
|   Other psychiatric disorders | 1.84 (1.59, 2.14) | < 0.01 | 1.15 (0.96, 1.38) | 0.13 |
|   Cognitive disorders | 1.30 (0.62, 2.75) | 0.49 | 0.87 (0.38, 2.02) | 0.75 |
| Health Service Utilization | | | | |
|   ED | | | | |
|     ED Non-MH visits | 3.97 (2.80, 5.63) | < 0.01 | 2.68 (1.85, 3.88) | < 0.01 |
|     Unplanned MH visits | 8.86 (7.65, 10.26) | < 0.01 | 2.06 (1.66, 2.54) | < 0.01 |
|     ED visits | 3.55 (3.07, 4.10) | < 0.01 | 1.06 (0.88, 1.28) | 0.54 |
|   Hospital | | | | |
|     Non-MH hospitalized | 7.26 (6.09, 8.65) | < 0.01 | 1.28 (1.0, 1.64) | 0.05 |
|     H visits | 3.69 (2.78, 4.91) | < 0.01 | 0.79 (0.57, 1.10) | 0.16 |
|     MH hospitalized | 7.70 (6.52, 9.11) | < 0.01 | 0.64 (0.50, 0.81) | < 0.01 |
| Clinic history visits | | | | |
|   Pharmacy | 2.06 (0.29, 14.77) | 0.47 | 2.39 (0.25, 22.48) | 0.45 |
|   Psychiatrist | 5.12 (4.41, 5.94) | < 0.01 | 1.49 (0.19, 1.85) | < 0.01 |
|   Other physicians | 1.31 (0.98, 1.75) | 0.07 | 1.18 (0.67, 2.10) | 0.57 |
|   Neurologist | 1.20 (0.96, 1.50) | 0.11 | 0.98 (0.75, 1.30) | 0.91 |
|   IM | 0.87 (0.75, 1.00) | 0.05 | 0.75 (0.63, 0.90) | < 0.01 |
|   GP | 0.85 (0.62, 1.17) | 0.33 | 0.43 (0.22, 0.83) | < 0.01 |
| Comorbidities | | | | |
|   Valvular disease | 1.73 (0.93, 3.23) | 0.09 | 1.97 (1.01, 3.85) | 0.05 |
|   Drug abuse | 12.41 (10.76, 14.33) | < 0.01 | 1.83 (1.46, 2.29) | < 0.01 |
|   Fluid and Electrolyte Disorders | 4.07 (3.45, 4.81) | < 0.01 | 1.51 (1.23, 1.85) | < 0.01 |
|   Hearing Impairment | 1.01 (0.72, 1.44) | 0.94 | 1.45 (1.01, 2.09) | 0.04 |
|   HIV/AIDS | 3.97 (1.77, 8.92) | < 0.01 | 1.43 (0.62, 3.29) | 0.41 |

Table 3    The performance metrics of all the predictive models. Abbreviations: LR, logistic regression; PT, power transformation for normalizing the data; ROTE, random over-sampling technique; J, Youden index; RF, random forest; SVM, support vector machine; ANN, artificial neural network; Sen, sensitivity; AUC, the area under the curve; Pre, precision; DF: dichotomous features; MF: multinomial features.

| | Primary Cohort with Initial homelessness outcome | | | Primary Cohort with Initial Police interaction outcome | | | Secondary Cohort with Initial homelessness outcome | | | Secondary Cohort with Initial Police interaction outcome | | |
|---|---|---|---|---|---|---|---|---|---|---|---|---|
| Logistic regression | Sen | AUC | Pre | Sen | AUC | Pre | Sen | AUC | Pre | Sen | AUC | Pre |
| Model 1 (LR model) | 0% | 50% | 0% | 0% | 50% | 0% | 4 % | 52 % | 39% | 4% | 52 % | 57 % |
| Model 2 (DF + LR model + J) | 82% | 83% | 4% | 78% | 82% | 0.3% | 90 % | 88 % | 9% | 82% | 84% | 8% |
| Model 3 (MF + LR model + PT + ROTE + J) | 85% | 84% | 3% | 88% | 81% | 1% | **91%** | **88%** | **9%** | **85%** | **85%** | **8%** |
| RF (MF+ PT +ROTE) | 76% | 83% | 6% | **65%** | 79% | 3% | 87% | 88% | 12% | 85% | 88% | 14% |
| XGBoost (MF + PT +ROTE) | 82% | 85% | 5% | **80%** | 75% | 2% | **91%** | **90%** | **12%** | 90% | 89% | 11% |

Table 4    The performance metrics of all the predictive LR models for homelessness and police interaction using dichotomous features. Abbreviations; MRLT: minimum data record length threshold that refers to the minimum duration of recorded data (in days) required for an individual to be included in the cohort or analysis; f1: F1-scoure; AUC: area under the curve; Sen: Sensitivity; Prec: Precision; Num-Out: Number of individuals with the related adverse outcome; Num-ind: number of individuals examined by the LR models.

| Homelessness | | | | | | | Police Interaction | | | | | | |
|---|---|---|---|---|---|---|---|---|---|---|---|---|---|
| f1 | AUC | Sen | Prec | MRLT | Num-Out | Num-ind | f1 | AUC | Sen | Prec | MRLT | Num-out | Num-ind |
| 0.17 | 0.88 | 0.90 | 0.09 | 360 | 3,814 | 236346 | 0.14 | 0.86 | 0.87 | 0.07 | 180 | 3,567 | 237,829 |
| 0.17 | 0.87 | 0.90 | 0.10 | 30 | 4,266 | 238823 | 0.11 | 0.85 | 0.88 | 0.06 | 720 | 2,894 | 232,869 |
| 0.17 | 0.87 | 0.90 | 0.09 | 0 | 4,417 | 240,218 | 0.14 | 0.84 | 0.82 | 0.08 | 360 | 3,401 | 236,498 |
| 0.15 | 0.87 | 0.90 | 0.08 | 180 | 4,034 | 237727 | 0.11 | 0.83 | 0.87 | 0.06 | 60 | 3,687 | 238,640 |
| 0.13 | 0.86 | 0.92 | 0.07 | 720 | 3,225 | 232639 | 0.11 | 0.83 | 0.86 | 0.06 | 0 | 3,837 | 240,218 |
| 0.16 | 0.86 | 0.89 | 0.09 | 60 | 4,209 | 238591 | 0.16 | 0.83 | 0.79 | 0.09 | 90 | 3,647 | 238,418 |
| 0.15 | 0.86 | 0.89 | 0.08 | 90 | 4,152 | 238355 | 0.12 | 0.82 | 0.84 | 0.07 | 30 | 3,720 | 238,852 |

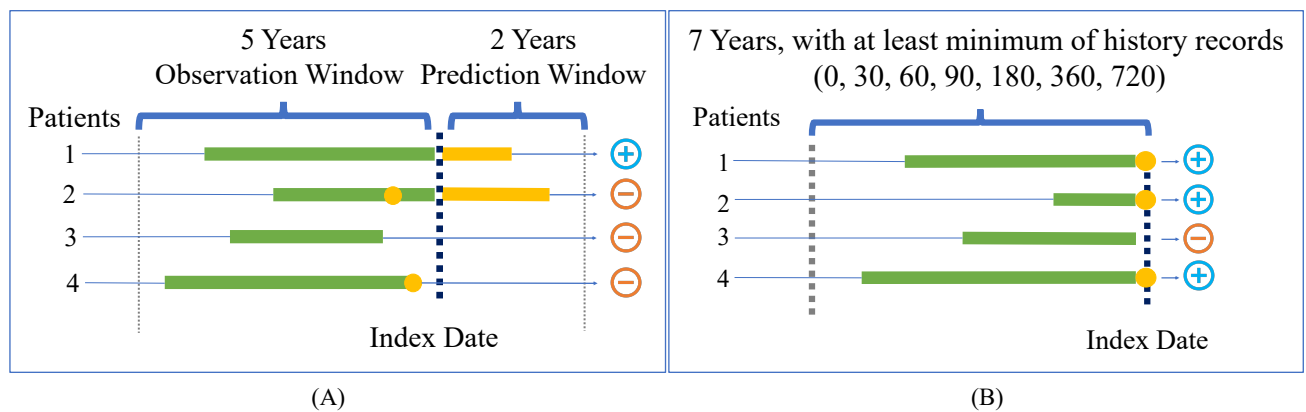

Figure 1    Framework for data extraction for predictive modeling tasks. (A) Primary

Cohort; Observation and prediction window with fixed size length and fixed index date (B) Secondary Cohort; Observation window with flexible size length for each patient; Index date is the first occurrence of outcomes.

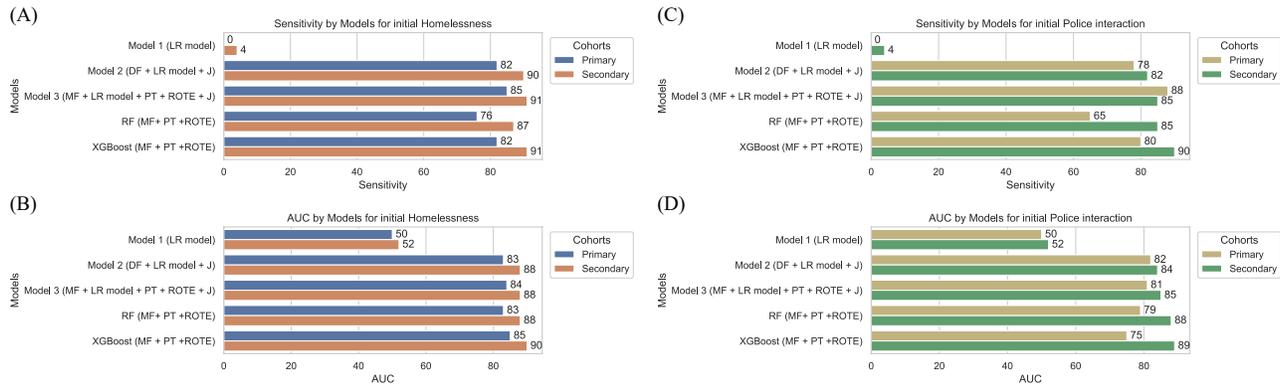

Figure 2    Provides a visualization of the results obtained from different models for each outcome. Subplot (A) displays the sensitivity gained by all models for the initial homelessness, while subplot (B) showcases the AUC values obtained by different models for the same outcome. Subplot (C) focuses on the sensitivity gained by all examined models for police interactions, and finally, subplot (D) presents the AUC values of all examined models for the initial police integration outcome.

*Supplementary Material*

# The Effect of Epidemiological Cohort Creation on the Machine Learning Prediction of Adverse Mental Health Outcomes Using Electronic Health Records Data

Faezehsadat Shahidi, M. Ethan MacDonald, Dallas Seitz, Geoffrey Messier [*]

**\* Correspondence:** Corresponding Author: gmessier@ucalgary.ca

## 1 Supplementary Figures and Tables

### 1.1 Supplementary Table 1: The hyperparameters for each ML model. Abbreviations: Num Exam; number of examinations for homelessness and police interaction (2*), RF; Random Forests, XGBoost; extreme gradient boosting

| Cohort | Model | Optimum hyperparameters | Experimented hyperparameters | Num Exam |
|---|---|---|---|---|
| 1 | RF | bootstrap= 'TRUE'<br>max_depth= 10<br>max_features= 'log2<br>min_samples_leaf= 8<br>min_samples_split= 10<br>n_estimators= 1000<br>criterion= 'entropy' | bootstrap= ('True')<br>max_depth= (10, 20, 30, 40, 50, 60, 70, 80, 90, 100)<br>max_features= ('sqrt', 'log2')<br>min_samples_leaf= (1, 2, 4, 8)<br>min_samples_split= (2, 5, 10)<br>n_estimators= (400, 600, 800, 1000, 1200, 1400, 1600, 1800, 2000)<br>criterion= ('gini', 'entropy') | 2* 4,320 |
| 1 | XGBoost | subsample=0.9<br>n_estimators=500<br>max_depth=3<br>learning_rate=0.01<br>colsample_bytree=0.7<br>colsample_bylevel=0.9 | subsample= (0.7, 0.8, 0.9)<br>n_estimators= (100, 500, 1000)<br>max_depth= (3, 5, 6, 10, 15, 20)<br>learning_rate= (0.01, 0.1, 0.2, 0.3, 0.4)<br>colsample_bytree= (0.4, 0.5, 0.6, 0.7, 0.8, 0.9)<br>colsample_bylevel= (0.4, 0.5, 0.6, 0.7, 0.8, 0.9) | 2* 9,720 |
| 2 | RF | bootstrap= 'TRUE'<br>max_depth= 10<br>max_features= 'log2<br>min_samples_leaf= 8<br>min_samples_split= 10<br>n_estimators= 600<br>criterion= 'entropy' | bootstrap= ('True')<br>max_depth= (10, 20, 30, 40, 50, 60, 70, 80, 90, 100)<br>max_features= ('sqrt', 'log2')<br>min_samples_leaf= (1, 2, 4, 8)<br>min_samples_split= (2, 5, 10)<br>n_estimators= (400, 600, 800, 1000, 1200, 1400, 1600, 1800, 2000)<br>criterion= ('gini', 'entropy') | 2* 4,320 |
| 2 | XGBoost | subsample=0.8<br>n_estimators=1000<br>max_depth=3<br>learning_rate=0.01<br>colsample_bytree=0.6<br>colsample_bylevel=0.9 | subsample= (0.7, 0.8, 0.9)<br>n_estimators= (100, 500, 1000)<br>max_depth= (3, 5, 6, 10, 15, 20)<br>learning_rate= (0.01, 0.1, 0.2, 0.3, 0.4)<br>colsample_bytree= (0.4, 0.5, 0.6, 0.7, 0.8, 0.9)<br>colsample_bylevel= (0.4, 0.5, 0.6, 0.7, 0.8, 0.9) | 2* 9,720 |

## 1.2 Supplementary Table 2: Data Sources and the Codes in Details

| Data Source | Codes | Details |
|---|---|---|
| 1. DAD | ICD-10-CA for comorbidities (79) | Type: DXCODE1 - DXCODE 25 |
| | ICD-10-CA for substance use disorder | Any hospitalization: F10.X—F19.X, F55.X, F63.X |
| | ICD-10-CA for mood disorder | F30.X, F31.X, F32.X, F33.X, F34.X, F38.X, F39.X, F53.0 |
| | ICD-10-CA for anxiety disorder | F40.X, F41.X, F42.X, F43.X, F48.8, F48.9 |
| | ICD-10-CA for psychotic disorder | F06.0, F06.1, F06.2, F06.0-2, F20.X, F22.X, F23.X, F24.X, F25.X, F26.X, F27.X, F28.X, F29.X, F22-F29, F53.1 |
| | ICD-10-CA for cognitive disorders | F00X, F01.X, F02.X, F03.X G30.X |
| | ICD-10-CA for Other psychiatric disorders | F06-F99 |
| | ICD-10-CA for Deliberate self-harm | X60-X84, Y10-Y19, Y28 when DX10CODE1 was not equal to F06-F99 |
| | ICD-10-CA for homelessness | Z590, Z591 |
| | ICD-10-CA for police interaction | Y350-Y357, Z650-Z653 |
| | MPSERV for the number of hospitalizations to a designated psychiatric facility | 64 |
| | MPSERV number hospitalizations to non-psychiatric facility | Not equal to 64 |
| 2. Claims | ICD-9-CM for comorbidities | HLTH_DX_ICD9x_CODE_1, HLTH_DX_ICD9x_CODE_2 HLTH_DX_ICD9x_CODE_3 |
| | ICD-9-CM for substance use disorder | 2 or more physician claims in 2 years with diagnosis: 291, 292, 303, 304, 305 |
| | ICD-9-CM for mood disorder | 2 physicians claim at least 30 days apart within 2 years with one or more of the diagnoses codes: 296, 311 |
| | ICD-9-CM for anxiety disorder | 2 physicians claim at least 30 days apart within 2 years with one or more of the diagnoses codes: 300, 308, 309 |
| | ICD-9-CM for psychotic disorder | 2 physicians claim at least 30 days apart within 2 years with one or more of the diagnoses codes: 295, 297, 298 |
| | ICD-9-CM for cognitive disorders | 3 physicians claim at least 30 days apart within 2 years with one or more of the diagnoses codes: 290, 331 |
| | ICD-9-CM for Other psychiatric disorders | 2 physicians claim at least 30 days apart within 2 years with one or more of the diagnoses codes: 290 – 319 |
| | ICD-9-CM for homelessness | V600, V601 |
| | ICD-9-CM for police interaction | e97.0-e97.6, e97.8, v62.5, z65.0-z65.3, E970-E976, E978, V625 |
| | PERS_CAPB_PRVD_SPEC_AD for Health Service Utilization | GP: family physician visits<br>NEUR: visits by neurologists<br>INMD: visits by general internal medicine<br>PSYC: visits by psychiatrists |
| 3. PIN | DINS codes for cognitive disorders | Donepezil, Galantamine, Rivastigmine or Memantine |
| 2- NACRS | ICD-10-CA for comorbidities (79) | Type: DXCODE1 - DXCODE 25 |
| | ICD-10-CA for a fall-related injury | W01, W10, W11, W12, W13, W06-W08, W09.01-W09.05, W09.08, W09.09, W14-W17, W00, W02.00-W02.05, W02.08, W03, W04, W05.00-W05.04, W05.08, W05.09, W18, W19 |
| | ICD-10-CA for substance use disorder | Any hospitalization: F10.X—F19.X, F55.X, F63.X |
| | ICD-10-CA for mood disorder | F30.X, F31.X, F32.X, F33.X, F34.X, F38.X, F39.X, F53.0 |
| | ICD-10-CA for anxiety disorder | F40.X, F41.X, F42.X, F43.X, F48.8, F48.9 |
| | ICD-10-CA for psychotic disorder | F06.0, F06.1, F06.2, F06.0-2, F20.X, F22.X, F23.X, F24.X, F25.X, F26.X, F27.X, F28.X, F29.X, F22-F29, F53.1 |
| | ICD-10-CA for cognitive disorders | F00X, F01.X, F02.X, F03.X G30.X |
| | ICD-10-CA for Other psychiatric disorders | F06-F99 |
| | ICD-10-CA for Deliberate self-harm | X60-X84, Y10-Y19, Y28 when DX10CODE1 was not equal to F06-F99 |
| | ICD-10-CA for homelessness | Z590, Z591 |
| | ICD-10-CA for police interaction | Y350-Y357, Z650-Z653 |